# Forecasting Future Humphrey Visual Fields Using Deep Learning

Joanne C. Wen MD,[1*] Cecilia S. Lee MD MS,[1*] Pearse A. Keane MD FRCOphth,[2] Sa Xiao PhD,[1] Yue Wu PhD,[1] Ariel Rokem PhD,[3] Philip P. Chen MD,[1] Aaron Y. Lee MD MSCI[1,3,4]

[*] These two authors contributed equally and should be considered co-first author.

[1] Department of Ophthalmology, University of Washington, Seattle WA

[2] NIHR Biomedical Research Centre for Ophthalmology at Moorfields Eye Hospital, Moorfields Eye Hospital NHS Foundation Trust and University College London (UCL) Institute of Ophthalmology, London, UK

[3] eScience Institute, University of Washington, Seattle WA

[4] Department of Ophthalmology, Puget Sound Veteran Affairs, Seattle Washington, USA

Corresponding author:

Aaron Y. Lee, MD MSCI
Assistant Professor
Department of Ophthalmology
University of Washington
Box 359608, 325 Ninth Avenue
Seattle WA 98104
Ph: (206) 543-7250
Email: leeay@uw.edu

**ABSTRACT**

Purpose: To determine if deep learning networks could be trained to forecast a future 24-2 Humphrey Visual Field (HVF).

Design: Retrospective database study.

Participants: All patients who obtained a HVF 24-2 at the University of Washington.

Methods: All datapoints from consecutive 24-2 HVFs from 1998 to 2018 were extracted from a University of Washington database. Ten-fold cross validation with a held out test set was used to develop the three main phases of model development: model architecture selection, dataset combination selection, and time-interval model training with transfer learning, to train a deep learning artificial neural network capable of generating a point-wise visual field prediction.

Main outcome measures: Mean absolute error (MAE) and difference in Mean Deviation (MD) between predicted and actual future HVF.

Results: More than 1.7 million perimetry points were extracted to the hundredth decibel from 32,443 24-2 HVFs. The best performing model with 20 million trainable parameters, CascadeNet-5, was selected. The overall MAE for the test set was 2.47 dB (95% CI: 2.45 dB to 2.48 dB). The 100 fully trained models were able to successfully predict progressive field loss in glaucomatous eyes up to 5.5 years in the future with a correlation of 0.92 between the MD of predicted and actual future HVF ($p < 2.2 \times 10^{-16}$) and an average difference of 0.41 dB.

Conclusions: Using unfiltered real-world datasets, deep learning networks show an impressive ability to not only learn spatio-temporal HVF changes but also to generate predictions for future HVFs up to 5.5 years, given only a single HVF.

**INTRODUCTION**

Glaucoma is a leading cause of blindness worldwide.[1] Of the 32.4 million blind people globally in 2010, an estimated 4.0 to 15.5% were from glaucoma.[2] Visual field (VF) testing using standard automated perimetry remains the gold standard for monitoring glaucomatous damage, and estimating future VF loss is integral to glaucoma management. With Humphrey standard automated perimetry, linear regression of global indices such as mean deviation (MD) and visual field index (VFI) are commonly used to assess glaucoma progression and risk of future progression.[3] However, as glaucomatous visual field changes are frequently focal, these global indices may fail to detect more subtle changes. Point-wise linear regression analyses were developed to evaluate changes at each test point and have been shown to identify glaucomatous progression earlier than global indices.[4] A number of linear and non-linear regression models have been described,[5–7] yet these models often assume a constant amount or rate of progression and typically evaluate progression rates of single visual field test points independent of adjacent test points. Furthermore, a number of VF tests are frequently required in these models to achieve reasonably accurate predictions.[7]

Recently, predictive machine learning methods have been developed with the advent of deep neural networks. Specifically the integration of "spatially aware" convolutional filters [8] paired with non-linear activation functions [9] has allowed computer vision researchers to achieve unprecedented accuracy in classification of real-world objects. These data-driven methods are able to directly ingest pixel intensities of a photograph and learn spatial features important for a given task. Ophthalmologic applications of deep learning include optical coherence tomography (OCT) image classification[10] and segmentation [11–13], fundus photo analysis,[14] and the diagnosis and classification of glaucoma based on disc photos, OCTs and VFs.[15–20] While most studies using deep learning have focused on classification, there has been limited application of deep learning in forecasting future findings. In addition to classification, deep learning algorithms can include segmentation, regression, localization, and generative models. The purpose of this study is to apply a deep learning generative model to predict future VFs with preserved spatial information using minimal baseline VF test input.

**METHODS**

This study was approved by the Institutional Review Board of the University of Washington and was in adherence with the tenets of the Declaration of Helsinki and the Health Insurance Portability and Accountability Act (HIPAA). All patients obtaining a Humphrey Visual Field (HVF) (Humphrey Visual Field Analyzer II-i, Carl Zeiss Meditec, Inc. Dublin, CA) using program 24-2 and the Swedish Interactive Thresholding Algorithm (SITA) between the years of 1998 and 2018 at the University of Washington were included in the study. Using an in-house generated, fully automated extraction algorithm that yields 54 perimetry point values to the hundredth decibel, consecutive extraction of HVF 24-2 was performed. Additional clinical variables such as age, gender, date of HVF, and eye tested were extracted and fully de-identified prior to subsequent analysis.

The resulting HVFs were then divided at the patient level into two parts: 80% for training and validation and 20% as a held-out test set. With the training and validation test, a total of 10 disjoint sets, divided again at the patient level, were created for 10-fold cross validation. For each set in the training, validation, and test sets, the HVFs were paired into every possible temporal combination, binned into 0.5 year intervals. Any combinations greater than 5.5 years and less than 0.75 years in interval were excluded. Three main phases of model development were then performed: model architecture selection, dataset combination selection, and time-interval model training with transfer learning. The final models weights were then frozen at each time-interval was used to assess the performance on the final held-out test set. The study design is depicted and summarized in Figure 1 and the evaluation strategy is shown in Supplementary Figure 1.

Model architecture selection was performed by testing 9 different structures (Table 1). Deep learning models were created such that they could take an input tensor with the HVF as a single 8x9 tensor and output a single 8x9 tensor representing the target HVF. The final layer activation was set to linear and the loss function for each model was masked mean absolute error (MAE). Each model was trained for 1000 epochs 10 times and validation accuracy was assessed at the end of each epoch using the interval data at one year (0.75 year to 1.25 years). The lowest validation accuracy was recorded for each model and 10-fold cross-validation. The best performing model was then carried forwarded to all the subsequent training.

Clinical dataset combination selection was done by taking the best performing model in the previous step and testing every possible combination of clinical predictor variables. Categorical variables such as eye and gender were appended in a one-hot vector format to the input tensor and continuous variables were encoded as a single additional tensor face with every cell encoded as the continuous value. Each data combination was trained for 1000 epochs 10 times and validation accuracy was assessed at the end of each epoch using the interval data at one year (0.75 year to 1.25 years). The lowest validation accuracy was recorded for each data combination and 10-fold cross-validation. The best performing data combination was then carried forwarded to the all the subsequent training.

After the best performing model and data combination was determined in the prior two phases, time intervals from 0.75 years to 5.5 years in 0.5 year intervals were trained. Transfer learning was performed with weights carried from the first temporal interval, and consecutive transfer learning with weights carried from the immediately prior time interval. Final accuracy and performance was assessed by predicting the final held out test set. The mean predictions from the 10 fold cross validation were used to generate each prediction.

All deep learning was performed with Keras (version 2.1.2, https://keras.io/) and Tensorflow (version 1.4.1, https://www.tensorflow.org/). Custom code was written in Python (version 2.7.12, https://www.python.org/) to scale and automate the training and all training was performed using a single server equipped with 8 x Nvidia Tesla P100 graphics processing units with NVIDA cuda (v8.0) and cu-dnn (v6.0) libraries (http://www.nvidia.com). Statistical analyses were performed using R (version 3.4.3, https://www.r-project.org/)

**RESULTS**

A total of 32,443 24-2 HVFs were extracted from 1998 to 2018 contiguously resulting in 1,751,922 perimetry points with hundredth decimal precision. All available HVFs were used and no filtering or data cleaning was performed prior to the training of the models. The HVFs were split into training, validation, and held-out test sets as described in Figure 1. Table 1 shows the baseline demographic factors of the study population.

Model development occurred in two distinct phases using HVF pairs from 0.75 months to 1.25 months. First, many different model architectures were explored (Table 2). For every model, 10-fold cross validation training was performed and the lowest validation MAE was recorded.

CascadeNet-5 (Supplementary Figure 2) resulted in the best network performance compared to the other network tested (Figure 2A).

The second phase of model development was to test giving the neural networks more clinical context. With the available predictors including age, gender, eye, and the HVF test number (ie the first, second or third, etc. VF test a patient has taken), all 16 possible combinations were trained and the lowest validation MAE was recorded. Consistently including age as an input into the deep learning model yielded superior performance (Figure 2B). Using CascadeNet-5 and age as an additional input parameter to the first HVF, continuous forward transfer learning was performed to train a separate model for each 0.5 year intervals. The average predictions from final trained models were then used the evaluate the held-out test set.

The mean deviations between the actual future HVF and the artificial intelligence (AI)-predicted HVF were statistically significantly correlated (Figure 2C, $r = 0.92$, Adjusted $R^2 = 0.84$, $p < 2.2 \times 10^{-16}$). Bland-Altman plot (Figure 2D) revealed a mean difference of +0.41 dB. The overall MAE (Figure 2E) and root mean squared error (RMSE) for the test set was 2.47 dB (95% CI: 2.45 dB to 2.48 dB) and 3.47 dB (95% CI: 3.45 dB to 3.49 dB), using the evaluation strategy shown in Supplementary Figure 2. Figure 3 shows representative predictions at different time points against the real HVFs for glaucoma of varying severity, as well as an example of a non-progressing non-glaucomatous hemifield VF defect. Figure 4 shows two examples of serial history in the test set where the same input HVF is used to predict multiple time points in the future.

**DISCUSSION**

Using more than 1.7 million perimetry points from 32,443 consecutive 24-2 HVFs performed during 20 years of routine clinical practice, deep learning was able to effectively predict the progression of VF defects seen in HVFs. The algorithm was able to generate predictions on HVFs with precise changes in the perimetry points occurring between 0.5 to 5 years. The only clinical factor that improved the AI's prediction was the patient's age. The overall MAE for the test set was 2.57 dB (95%CI: 2.56 dB to 2.59 dB) and the RMSE was 3.47 (95% CI: 3.45 dB to 3.49 dB).

Prior algorithms for generating predictions of HVFs have centered around fitting regression models for consecutive HVFs and extrapolating the next HVF. Although easy to interpret, simple linear regression analyses such as mean deviation, pattern standard deviation or visual field index do not take into account the spatial nature of visual field loss in glaucoma.[21] Pointwise, exponential regression models have been developed to characterize fast or slow progression rate in VF loss better than linear models.[22] However, these models assume a constant rate of VF deterioration and fail to address varying rates of glaucomatous decay over time.[5] Multiple model approaches that include exponential and logistic function appear to predict future progressions better.[6] Chen et al. reported average RMSEs of 2.925 and 3.056 for logistic and exponential functions, respectively.[6] More recently, a pointwise sigmoid regression model with a mean RMSE of 4.1 has been described that captures the varying rates of glaucomatous deterioration over time and characterizes both early and late stages of glaucoma.[5]. Another approach has been in the Bayesian setting to incorporate the spatial data in the VF. Long-term variability in HVF has been one of the major limitations in assessing progression in HVF, and several models have been described to take into account of spatio-temporal correlations in the perimetry data.[23–26] However, the majority of studies were limited by small sample size (in the hundreds), data cleaning of the training sets, and evaluation/validation sets with stringent criteria.

Current models that predict future VFs require many VF tests obtained over relatively long periods of time. Using a point-wise ordinary least squares linear regression, Taketani et al. found that the minimum absolute prediction error of 2.4 ± 0.9 dB was achieved using the maximally available number of 14 VFs.[7] As these VFs are acquired approximately every 6 months, many years of testing would be needed to accurately predict future VFs. To decrease the number of necessary VFs, Fujino et al. applied a least absolute shrinkage and selection operator regression and reported a prediction error of 2.0 ± 2.2 dB using only 3 VFs.[27] However, even with this model, over a year would pass before an accurate prediction could be made. The most remarkable aspect of our approach is that by using deep learning models trained on the temporal history for a large group of patients, we are able to take single HVFs and predict point-wise HVFs at half-year time intervals, up to five years later. To our knowledge, no prior algorithm has been shown to reliably predict the VF changes based on a single HVF such as ours, to a comparable level of accuracy.

In general, the predicted HVFs tended to be slightly less progressed than the actual HVFs, as seen in the Bland-Altman plot (FIgure 2D), whereby the mean difference in MD between predicted and actual HVFs was 0.41 dB. This may be because the dataset used to train our algorithm was unfiltered. Our dataset included non-glaucomatous VFs and VFs after ocular surgeries that may increase perimetry sensitivity, such as cataract extraction. The majority of studies on forecasting VFs and estimating VF progression utilize highly filtered datasets with strict inclusion and exclusion criteria. Most require a certain number of VF tests per study subject of a certain reliability or a known diagnosis of glaucoma before inclusion. Notably, our model was developed using completely unfiltered data and included every single VF test performed at our institution regardless of diagnosis or test reliability, and yet the algorithm was still able to make highly accurate predictions. Even in cases of non-glaucomatous VF changes, such as a hemifield defect from a cerebrovascular accident (Figure 3), the program forecasts an appropriate future VF, which in this example is a stable VF defect. Conversely, the VF for moderate glaucoma in Figure 3 has the appearance of a quadrantanopia on cursory visual inspection, yet the program was able to accurately predict it would progress in a glaucomatous manner. The algorithm would likely perform even better were the program trained and evaluated on a rigorously filtered dataset.

Previous applications of deep learning in glaucoma have been limited to classification rather than forecasting and included analysis of disc photos,[17,19,28] OCT images[18,29] and VFs.[15] Recently, Li et al. used a deep learning system to classify over 48,000 optic disc photos for referable glaucomatous optic neuropathy and reported an area under the receiver operating characteristic curve (AUROC) of 0.986 with a 95.6% sensitivity and 92.0% specificity.[19] Using a hybrid deep learning method to analyze 102 single wide-field OCT scans, Muhammad et al. found that their protocol outperformed standard OCT and VF clinical metrics in differentiating suspects from early glaucomatous eyes.[18] Asaoka et al. used a multilayer, feed-forward neural network to detect preperimetric glaucoma VFs (AUC 92.6%, 95% CI 89.8-95.4) using 171 preperimetric VFs and 108 control VFs.[15] The deep learning classifier performed significantly better than other machine learning algorithms in distinguishing preperimetric glaucoma from healthy VFs. Unlike these previous classification studies, our study demonstrates a novel application of deep learning to build generative models in glaucoma.

Ideally, forecasted VFs would guide clinical decision-making and disease management. Kalman Filter models have recently been developed to determine the best VF testing frequency to most efficiently detect glaucoma progression and also to forecast glaucoma progression trajectories at different target intraocular pressures.[30,31] Our current model incorporates very little clinical data, although the predictive accuracy was already improved after including subject age. Future research will focus on developing models that incorporate additional clinical parameters, such as IOP, medication and surgical intervention, to assess their effects on forecasted VFs and progression, which may aid in clinical decision making. Additionally, our model could be used to determine how HVFs would change depending on the intervention selected for each individual patient. This type of AI application may help enable precision medicine for glaucoma management in the future.

Our study has several limitations. First, this study included data from a single academic center, thus our AI-algorithm may not be widely applicable. However, we included all available HVFs over a twenty year period, including numerous providers of different subspecialties, with no filtering. Future work may include training with datasets from multiple centers and testing in a geographically separated cohort. Second, our test set did not include any pairs of HVFs that were beyond 5.5 years of duration. It is possible that longitudinal data of longer duration (i.e. >5.5 years) may have included more HVFs with progression to advanced glaucomatous loss resulting in substantial change in the decay rate. Under such circumstances, our algorithm may have performed differently from our current results. Finally, the AI-algorithm was not trained as a classification tool; rather, it was constructed to generate prediction maps for future HVFs that incorporate both spatial and temporal data, a much more difficult task for artificial intelligence than simply classifying a subject as progressed or stable.

In conclusion, we have developed a novel deep learning algorithm that is able to forecast HVFs up to 5 years from a single HVF as input with remarkable accuracy. The ability to quickly anticipate future glaucomatous progression may prevent unnecessary functional loss that can occur with the current practice of multiple confirmatory tests. With further incorporation of clinical data such as IOP, medication and surgical history, our AI-model may assist in clinical decision-making in the future and allow development of a personalized treatment regimen for each patient.

**TABLES**

**TABLE 1: Baseline demographics at patient eye level.**

|  | Training / Validation Set | Held-out Test Set | Total |
|---|---|---|---|
| Patients (n) | 3,972 | 340 | 4,312 |
| Eyes (n) | 6,536 | 1,737 | 8,273 |
| HVFs (n) | 25,723 | 6,720 | 32,443 |
| Mean age at first HVF (SD) | 62.0 (14.8) | 61.8 (15.1) | 61.9 (14.8) |
| Gender |  |  |  |
|   Male, n (%) | 3,026 (46.3%) | 768 (44.2%) | 3,789 (45.8%) |
|   Female, n (%) | 3,510 (53.7%) | 969 (55.8%) | 4,484 (54.2%) |
| Eye |  |  |  |
|   Right, n (%) | 3,281 (50.2%) | 869 (50.0%) | 4,153 (50.2%) |
|   Left, n (%) | 3,255 (49.8%) | 868 (50.0%) | 4,120 (49.8%) |
| Mean number of HVFs (SD) | 3.7 (1.9) | 3.6 (1.8) | 3.6 (1.9) |
| Mean follow-up between first and last HVFs in years (SD) | 3.5 (2.8) | 3.4 (2.8) | 3.5 (2.8) |
| Average mean deviation (SD) | -6.73 (6.26) | -6.70 (6.10) | -6.73 (6.23) |

**TABLE 2: Different model architectures trained for 1,000 epochs.**

|  | Layers (n) | Batch Size | Optimizer | Space Complexity (MB) | Parameters |
|---|---|---|---|---|---|
| Fully Connected | 2 | 32 | Adam $1\times10^{-3}$ | 4 | 336,968 |
| FullBN-3 | 10 | 32 | Adam $1\times10^{-3}$ | 23 | 1,921,795 |
| FullBN-5 | 16 | 32 | Adam $1\times10^{-3}$ | 40 | 3,472,771 |
| FullBN-7 | 22 | 32 | Adam $1\times10^{-3}$ | 58 | 5,023,747 |
| Residual-3 | 12 | 32 | Adam $1\times10^{-3}$ | 27 | 2,332,163 |
| Residual-5 | 18 | 32 | Adam $1\times10^{-3}$ | 45 | 3,883,139 |
| Residual-7 | 24 | 32 | Adam $1\times10^{-3}$ | 63 | 5,434,115 |
| Cascade-3 | 10 | 32 | Adam $1\times10^{-3}$ | 81 | 6,992,086 |
| Cascade-5 | 16 | 32 | Adam $1\times10^{-3}$ | 238 | 20,694,754 |

**FIGURES**

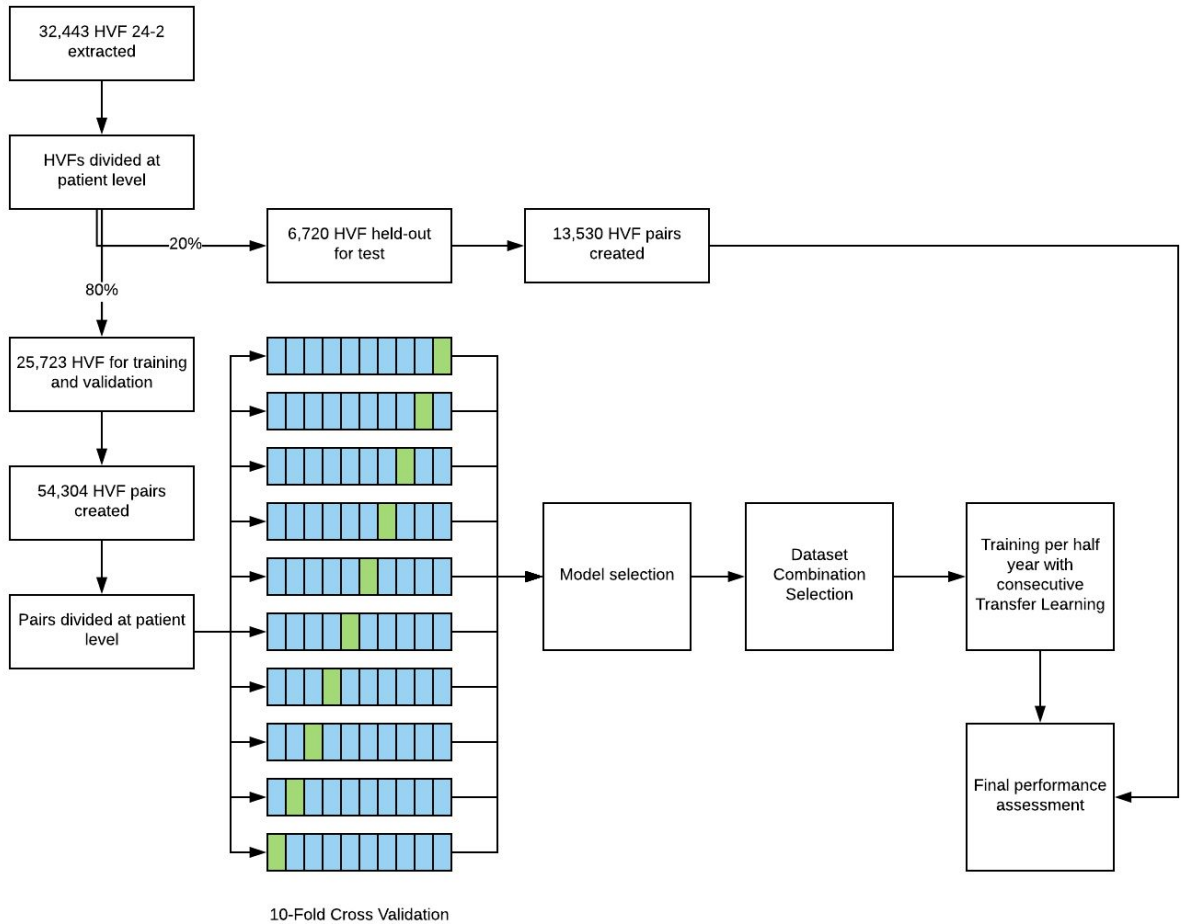

**Figure 1: Flowchart of study design.** A combination of cross validation and held out test set was used for the development and evaluation of the models.

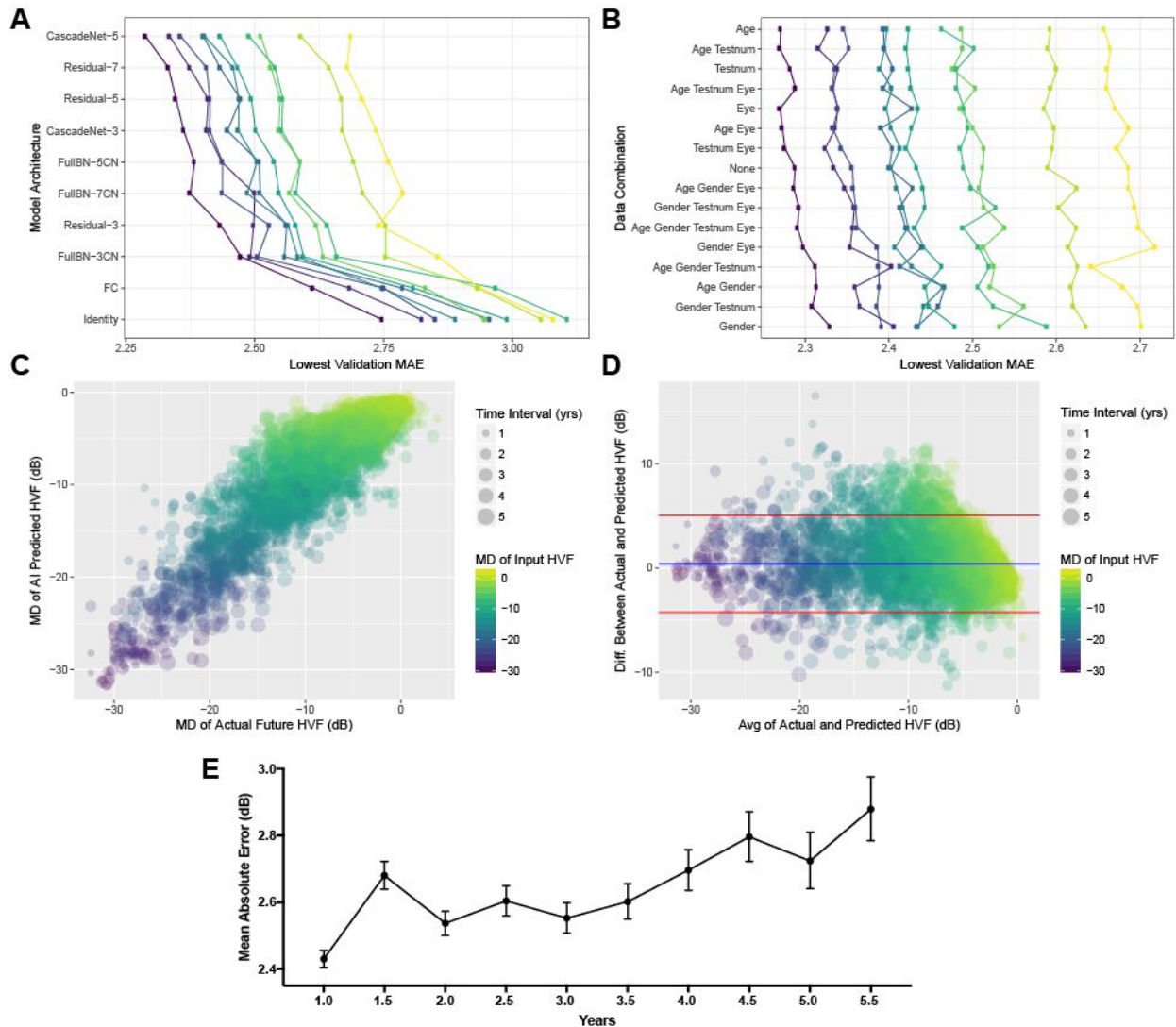

**Figure 2: Model development and evaluation.** A) Model architectures tested with lowest validation Mean Absolute Error (MAE) shown with each colored line representing one of the ten fold internal cross validation datasets. B) Data combinations tested with every possible combination of age, gender, eye, and test number (HVF #)**.** Model evaluation using held out test set with scatter plot (C) and Bland-Altman plot (D) between Mean Deviation (MD) of the AI predicted and actual future HVFs, (r = 0.92, Adjusted $R^2$ = 0.84, p < 2.2 x $10^{-16}$), color shaded by MD of input (earlier) HVF and sized by time interval. The blue and red lines in (C) represent the mean and 95% CI, respectively. E) Average MAE for each time interval of held-out test set with 95 % confidence intervals as error bars.

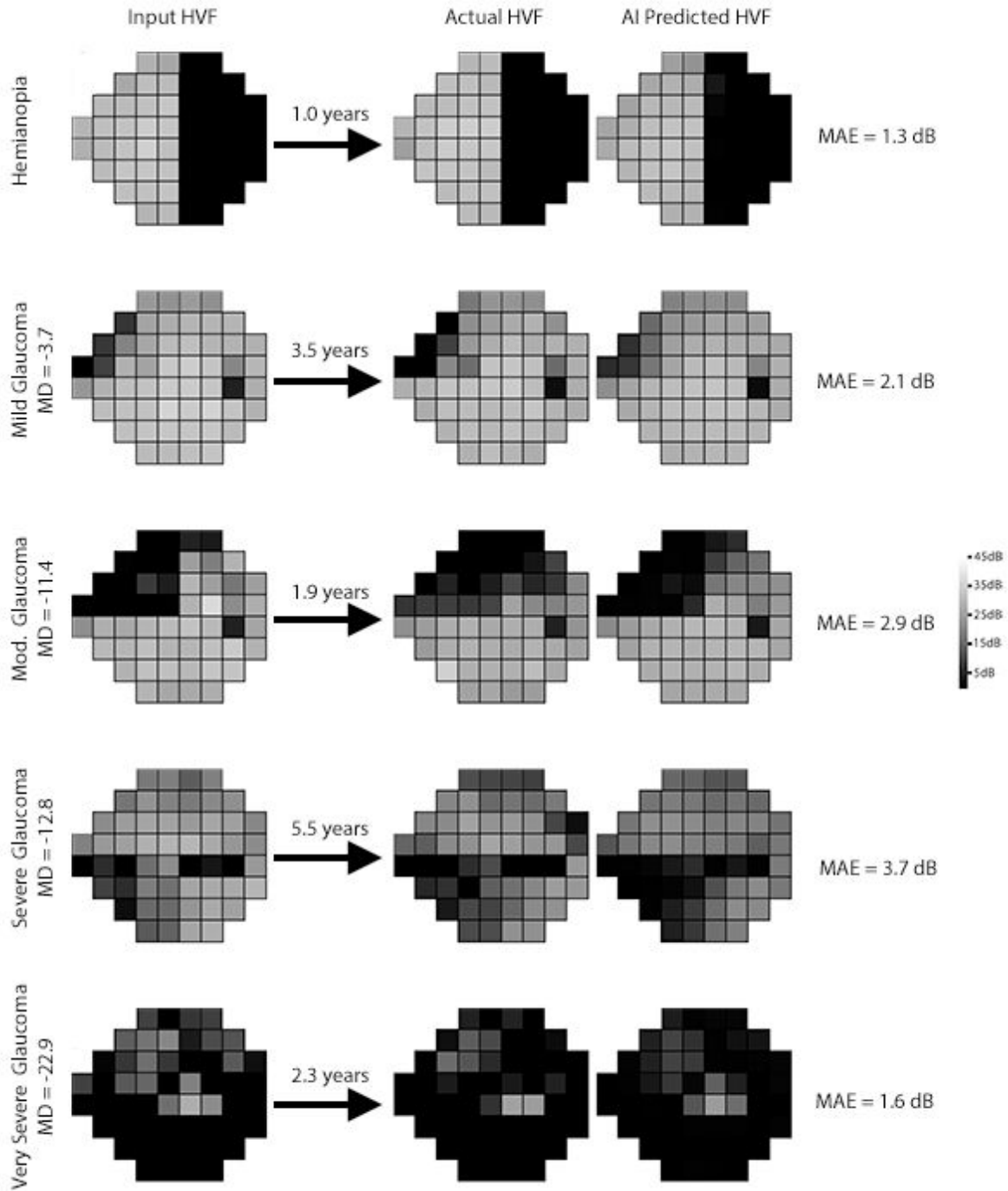

**Figure 3: Representative examples from the held-out test set, comparing actual and artificial intelligence (AI) predicted Humphrey Visual Field (HVF).** The input HVF is used for predicting HVF at the respective designated time points. A variety of different starting HVFs are shown ranging from a hemifield defect to very severe glaucoma with the corresponding Mean Absolute Error (MAE).

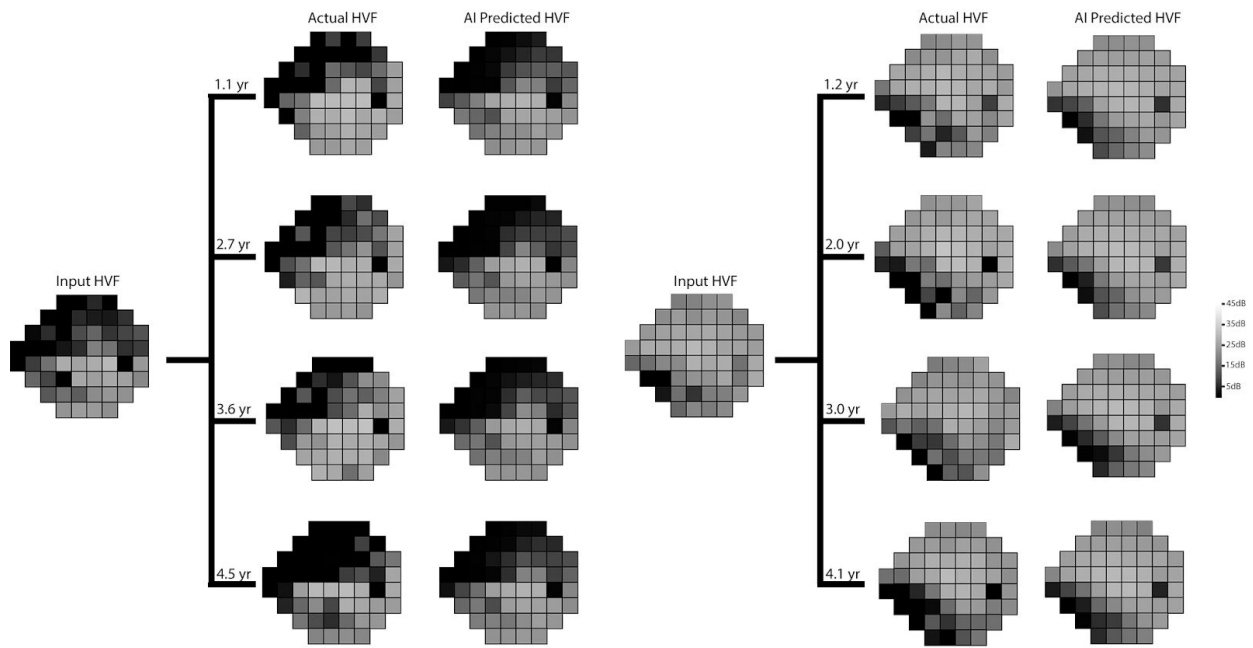

**Figure 4: Serial comparisons between actual and artificial intelligence (AI) predicted Humphrey Visual Field (HVF).** The single input HVF is used for multiple predictions at different time points. Mean absolute errors for the left panel from top to bottom were 3.8, 3.8, 5.1, and 4.0 respectively, and for the right panel from top to bottom were 2.3, 2.5, 3.2, and 3.2, respectively.

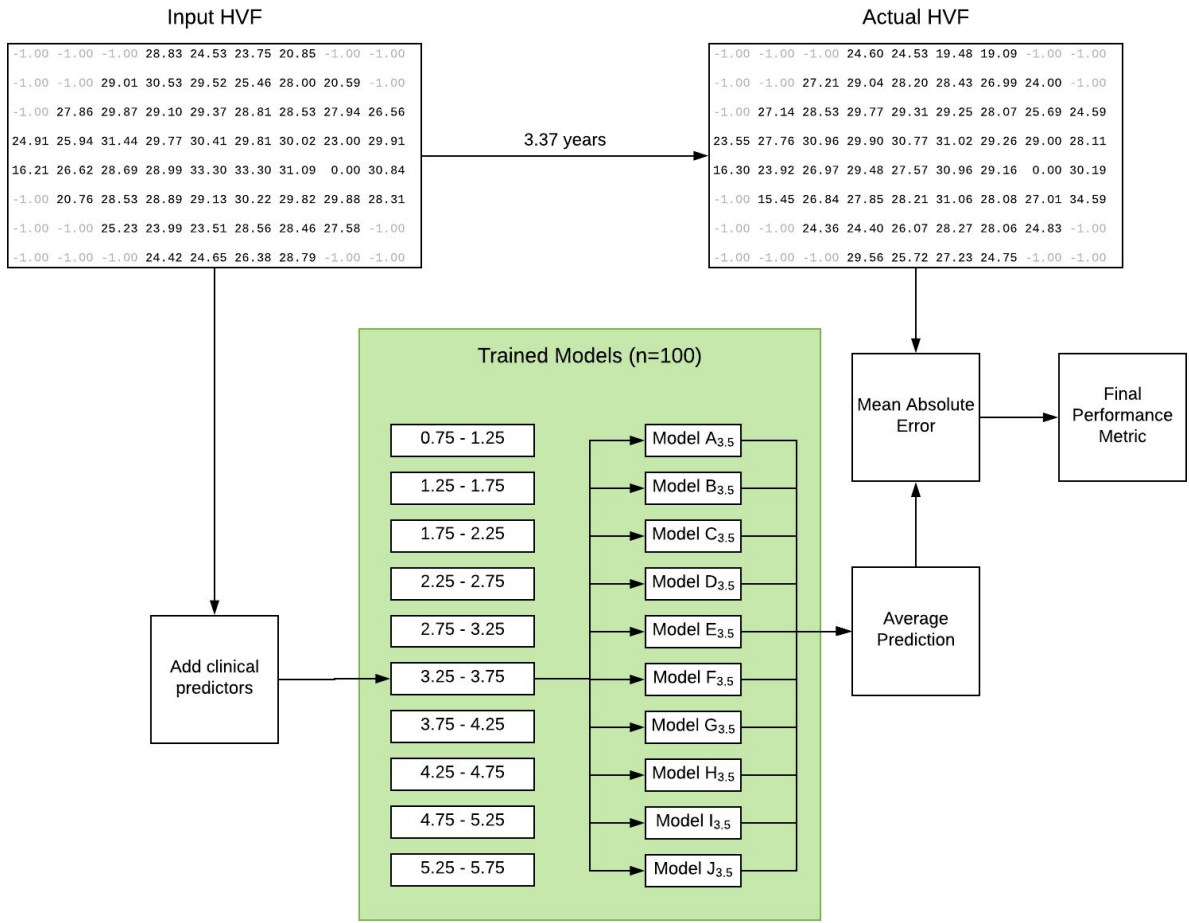

**Supplementary Figure 1: Evaluation strategy for test set after models training.** For each of the ten time interval, there are 10 trained models, one for each cross validation set. Each HVF prediction is then averaged together and the final prediction is compared with the actual HVF.

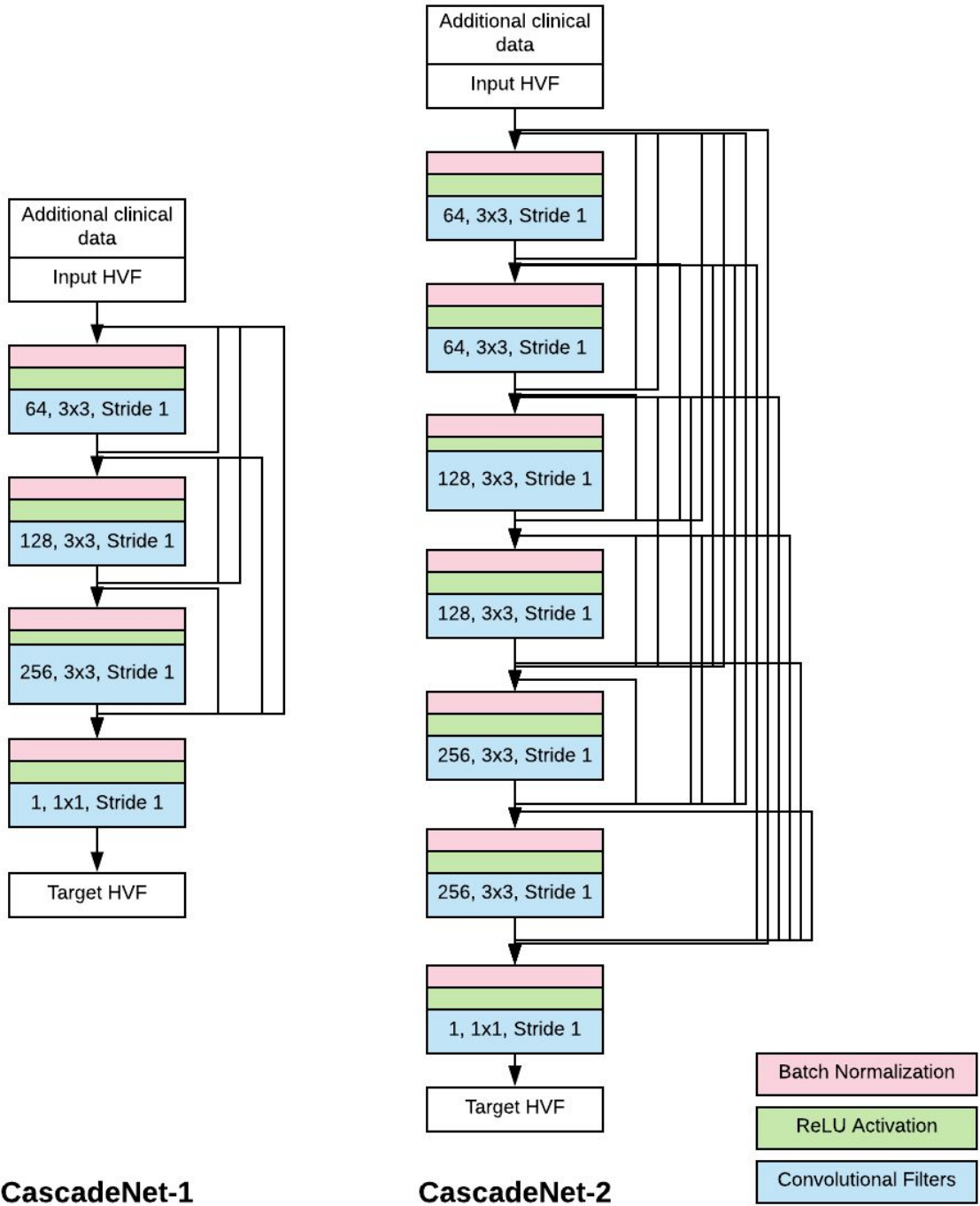

**Supplementary Figure 2: Simplified schematic of the CascadeNet architecture.**
CascadeNet-1 has 1 copy of convolutional (CN) layers of 64, 128, and 256 filters. CascadeNet-2 has 2 copies of CN layers of 64, 128, and 256 filters. Each black arrow represents a copy-concatenation of the output tensors "cascaded" forward to every possible neural network block. The final model, CascadeNet-5 has 5 copies of CN layers with 64, 128, and 256 filters.

**ACKNOWLEDGEMENTS**

We would like to acknowledge the Glaucoma and Neuro-Ophthalmology departments at the University of Washington from which the majority of these visual fields were derived. We would also like to acknowledge the NVIDIA Corporation for their generous donation of graphics cards for the development of artificial intelligence algorithms. This work was supported by the following grants: the National Eye Institute K23EY024921 (CSL), unrestricted research grant from Research to Prevent Blindness (CSL, AYL, SX, YW), The Gordon & Betty Moore Foundation (AR), and The Alfred P. Sloan Foundation (AR).